# Brain MRI Image Super Resolution using Phase Stretch Transform and Transfer Learning

Sifeng He[1,2] and Bahram Jalali[1,3,4]

1.Department of Electrical and Computer Engineering, UCLA, 2. Department of Precision Instrument, Tsinghua University

3.Department of BioEngineering, UCLA, 4. Department of Surgery, UCLA, USA

Author e-mail address: hesifeng@ucla.edu

**Abstract:** A hallucination-free and computationally efficient algorithm for enhancing the resolution of brain MRI images is demonstrated.

## 1. Introduction

Medical imaging is fundamentally challenging due to absorption and scattering in tissues and by the need to minimize illumination of the patient with harmful radiation. Imaging modalities also suffer from low spatial resolution, limited dynamic range and low contrast. These predicaments have fueled interest in enhancing medical images using digital post processing [1]. Recent progress in image super resolution using machine learning and in particular convolutional neural networks (CNNs) may offer new possibilities for improving the quality of medical images. However, the tendency of CNNs to hallucinate image details is detrimental for medical images as it may lead to false diagnostics. Also, these techniques require prohibitively large computational resource, a problem that is exacerbated by the large size of medical images. This has motivated us to investigate hallucination-free and computationally efficient algorithms for enhancement of medical images.

## 2. Algorithm

Our proposed approach combines the advantages of RAISR upscaling method [2] and phase stretch transform (PST) algorithm [3]. RAISR algorithm is developed by Google for improving conventional photography images [2]. We begin by dividing the input image into patches then cluster the patches based on gradient features. These features include gradient angle, strength and coherence used in the original in RAISR algorithm plus a feature obtained by applying the phase stretch transform (PST) [3] to each patch. PST is a physics inspired algorithm that emulates dispersive propagation and coherent detection. PST feature is obtained by applying the following algorithm [3,4]:

$$I_{PST}(x,y) = \sphericalangle \langle IFFT2\{\widetilde{K}[u,v] \cdot FFT2\{I(x,y)\}\}\rangle, \qquad (1)$$

where $I(x,y)$ is the input image and $\widetilde{K}[u,v]$ is a nonlinear frequency dependent phase.

    Therefore, the extracted feature from each patch contains four values: gradient strength, gradient coherence, gradient angle and PST feature. From the collected LR patches, we can extract a large set of patch features from them. Then the patch features can be clustered into $K$ buckets and these $K$ cluster centers can be viewed as anchor points to represent the feature space of natural image patches ($K$ is the cluster number). After constructing the feature space of LR image patches, all the LR patches in the training image dataset can be labelled by searching their feature nearest neighbor over learned dictionary. Therefore, the feature space is split into $K$ subspace. For each subspace, an appropriate filter connecting the high and low resolution patches is constructed during the training step using by solving a least-squares minimization problem:

$$\min_{h_q} \|A_q h_q - b_q\|_2^2, \qquad (2)$$

where $A_q$ and $b_q$ are the patches and pixels that belong to the $q$-th subspace. Hence, the filters for different kinds of patch features are obtained, leading to the local adaptivity for the next step upscaling. To conclude the training stage, we split the LR patch feature space into numerous subspaces and learn a separate mapping from LR to HR versions as priors for each subspace.

Given a LR test image, each LR patch $A_i$ is separated to compute the LR features and search for the closest cluster center. According to the cluster center with the index assumption of $q$, we apply the learned filter $h_q$ of the corresponding subspace to compute the corresponding HR pixel value $b_i$:

$$b_i = A_i \cdot h_q \qquad (3)$$

Then the pixels obtained from Eq.3 are combined together to construct the upscaled high resolution image.

## 3. Performance

For medical images, it is always difficult to acquire large training dataset. To deal with this problem, we implement transfer learning [5]. We train our model with natural (non-medical) images where large data sets exist, and apply them for medical image enhancement. Here, we name our proposed algorithm as Phase Stretch Anchored Regression (PhSAR) that employs PST to supplement features extracted by conventional techniques. Our proposed algorithm is compared with other baseline methods, including RAISR [2], A+ [6], ANR, GR [8] and SRCNN [7]. A quantitative comparison result for 2×, 3× and 4× upscaling is given in Fig.1. The implemented codes of RAISR algorithm can be referred from [9]. The difference in performance can be attributed to the transferability of learning from natural images to MRI images. It can be observed that both RAISR and our proposed method have good transferability. SRCNN has the worst transferability because it creates artificial details. These artifacts may have been presented in natural images used to train it, but that do not belong in the MRI image (image hallucination). This is expected because transferability of features learned by neural network decreases as the distance between the base task and target task increases. A+ algorithm performs slightly better than SRCNN. The observed difference in performance between our algorithm with A+ may be attributed to the difference between direct mapping of LR and HR images in our case vs. mapping of LR features to HR images in A+. Moreover, the average PSNR of PhSAR is about 0.3dB higher than RAISR algorithm in different upscaling factors. For the execution time, PhSAR can be demonstrated with best restoration performance without sacrificing the computational complexity, while the running time is only slightly slower than RAISR (the lack of clustering and hence nearest neighbor search in RAISR makes it slightly faster at the expense of lower PSNR).

In the case of upscaling factor 3, a visual comparison between different methods is provided by Fig.2. As can be seen, all the super resolution algorithms show significant sharpness improvements compared with cheap upscaling (bicubic interpolation). But among them, only SRCNN and PhSAR results have smooth edge, which can be clearly observed from the zoomed-in images of right side with red borders. However, after close examination, we find that SRCNN hallucinates artificial details, which makes PSNR even lower than other algorithms. For instance, inside the red circle in left zoomed-in images in Fig.2(h), SRCNN generates too sharp edge that is not consistent with original reference image. The outside edge is much narrower than the ground truth and other methods. This hallucination is more visually significant at high upscaling factor, shown in Fig.2(i) with upscaling factor of 4. Both the outside edge and inside edge are too sharped with distortion, while this does not happen on PhSAR or any other filter learning algorithm. This hallucination is detrimental for medical application while it may lead to false diagnosis. With improved sharpness and no hallucination, our proposed PhSAR algorithm produces the highest PSNR performance.

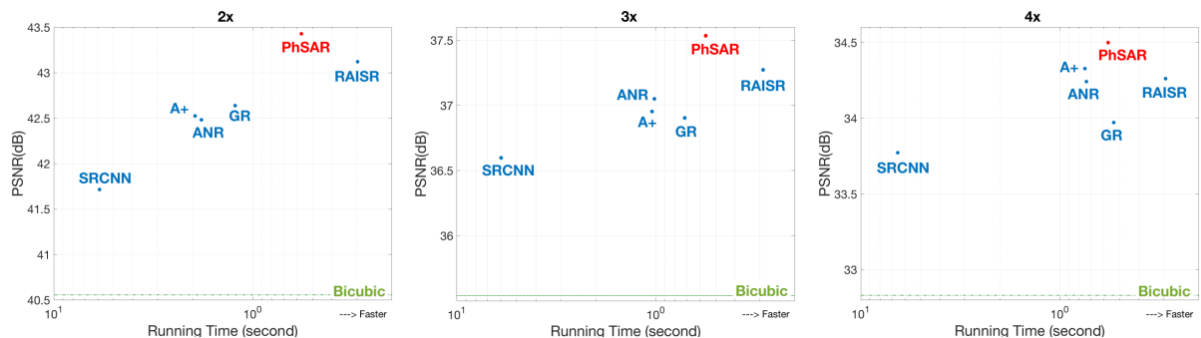

Fig.1 Quantitative comparison of the super resolution performance between different algorithms

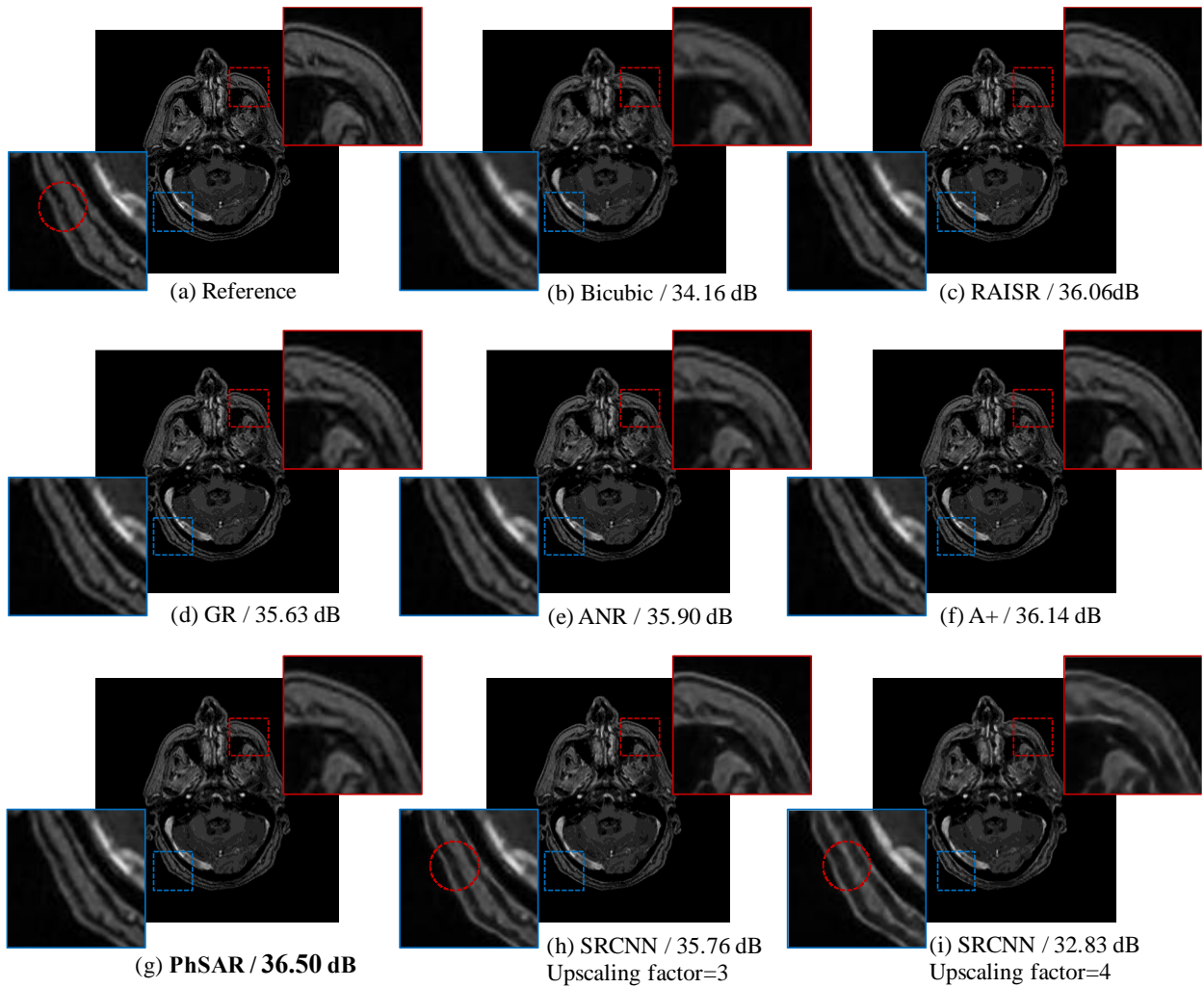

Fig.2 The medical image enhancement comparison of different methods in the upscaling factor of 3 (except the last figure).

## 4. Conclusion

We have proposed and demonstrated a learning based approach to enhance medical images. Different from the state-of-the-art neural network algorithm, our proposed algorithm is based on locally adaptive learned filtering leading to no hallucination. We train our model using natural images and then apply it to medical images to deal with the lack of large labeled image data sets. With the advantages of no hallucination, high restoration performance and low computational cost, our proposed algorithm provides an excellent solution for medical image super resolution.